\documentclass[runningheads]{llncs}
\usepackage[T1]{fontenc}
\usepackage{booktabs}
\usepackage{multirow} 
\usepackage{amsmath}
\usepackage{amssymb}
\usepackage{amsfonts}
\usepackage{amsmath}
\usepackage{marvosym}
\usepackage{graphicx, verbatim, subcaption}
\begin{document}
\title{RePCM: Region-Specific and Phenotype-Adaptive Bi-Ventricular Cardiac Motion Synthesis\thanks{This is the submitted version of the paper, before peer review. The final authenticated version will be available online at SpringerLink.}}
\titlerunning{RePCM}
%Region-Aware Synthesis of Phenotype-Adaptive Bi-Ventricular Cardiac Dynamics

\author{Xuan Yang\inst{1} \and
Xiaohan Yuan\inst{1,2}  \and
Hao Li\inst{1} \and
Lingyu Chen\inst{3} \and
Yanan Liu\inst{1,4} \and
Lei Li\inst{1}(\textsuperscript{\Letter})}
\authorrunning{X. Yang et al.}
\institute{
School of Biomedical Engineering, National University of Singapore, Singapore\\
\email{lei.li@nus.edu.sg}
\and
School of Automation, Southeast University, Nanjing, China
\and
School of Computer Science and Technology, Nanjing University of Aeronautics and Astronautics, Nanjing, China
\and
School of Information Science and Engineering, Yunnan University, Kunming, China
}

\maketitle              % typeset the header of the contribution
\begin{abstract}
Cardiac motion over a cardiac cycle is crucial for quantifying regional function and is strongly affected by cardiovascular diseases.  Since temporally dense mesh sequences are difficult to obtain in practice, we focus on leveraging the more accessible end-diastolic frame to infer a full-cycle sequence. Due to strong regional and disease-specific differences, traditional methods often oversmooth the data by relying on generative models that are optimized for global patterns. To address this problem, we propose Region-Aware and Phenotype-Adaptive Bi-Ventricular Cardiac Motion Synthesis (RePCM) for single frame Bi-ventricular mesh motion completion. In Stage I, a reconstruction network learns vertex wise motion descriptors and clustering yields a data driven functional partition, providing an explicit motion derived region structure. In Stage II, a Region-Specific Injection Module enforces masked, synchronized region exchange within a conditional VAE, preserving localized specific dynamics and restricting cross-region mixing. A Phenotype-Adaptive Mixture-of-Experts prior conditioned on ED shape uses anatomy-guided cues to model latent motion trends and capture inter-disease variability. Experiments on three datasets covering different cardiovascular diseases show consistent gains in geometric and functional metrics and improved preservation of region specific dynamics.

% Authors must provide keywords and are not allowed to remove this Keyword section.

\end{abstract}

\section{Introduction}
During the cardiac cycle, the heart undergoes complex three-dimensional deformations, with different chambers and anatomical regions exhibiting distinct contraction and relaxation patterns \cite{intro[2]}. These spatio-temporal motion characteristics are closely associated with a wide spectrum of cardiovascular diseases \cite{intro[1]}. Accurate modeling of such dynamics is therefore essential for regional functional assessment, understanding cardiac remodeling, and building individualized computational cardiac models. However, learning physiologically consistent cardiac dynamics remains challenging. Cardiac morphology and motion are intrinsically coupled, while different disease populations exhibit diverse motion patterns \cite{intro[3]}. In addition, clinical image sequences often suffer from irregular temporal sampling or missing frames, making it difficult to recover a complete cardiac cycle and maintain cross-phase consistency without densely observed sequences \cite{intro[4]}.

Early cardiac generative modeling largely relied on template registration and statistical shape models (SSM), which establish population-level comparisons through consistent parameterization \cite{SSM}. Although SSM offers good interpretability, the linear shape space and static modeling assumptions limit its ability to capture complex nonlinear deformations and temporal dynamics. More recently, implicit representations have been explored for complex anatomy \cite{SDF4CHD,STNDF}. SDF4CHD uses signed distance fields to accommodate congenital topological variations \cite{SDF4CHD}, and ST-NDF extends neural distance fields to spatiotemporal generation from a single frame \cite{STNDF}. However, implicit methods often lack stable topology and explicit point-wise correspondence, making them less suitable for fine-grained mesh-based dynamic analysis. In contrast, VAE-based frameworks generate explicit, vertex-aligned meshes, enabling reliable regional tracking and functional measurements. CHeart \cite{CHEART} performs conditional spatiotemporal generation using voxel labels and clinical variables, yet spatial and temporal variations are mainly captured in a shared latent space. MeshHeart \cite{MeshHeart} learns a canonical BiV 3D+t distribution for phenotypic analysis but focuses largely on healthy cohorts. CardioSynth4D \cite{Cadio4d} improves controllability via shape-motion disentanglement, but still lacks explicit mechanisms for fine-grained regional motion modeling.

To address these limitations, we propose \textbf{Re}gion-Aware and \textbf{P}henotype-Adaptive Bi-Ventricular \textbf{C}ardiac \textbf{M}otion Synthesis (\textbf{RePCM}), a motion generative  framework of oriented towards unified topological bi-ventricular meshes. This method generates full-cycle sequences using a single-phase end-diastolic (ED) mesh as input, simultaneously modeling regional motion constraints and cross-disease dynamic differences during the generation process. Specifically, we incorporate region-aware motion priors to preserve localized dynamics and improve cross-phase consistency, reducing over-smoothing across functionally different regions. In addition, we employ a shape-conditioned, phenotype-adaptive mixture-of-experts (MoE) prior to capture disease-specific motion variability by leveraging the intrinsic coupling between morphology and motion.

\section{Method}
%总述:强调多疾病 区域感知

\begin{figure*}[t]
  \centering
  \includegraphics[width=\textwidth]{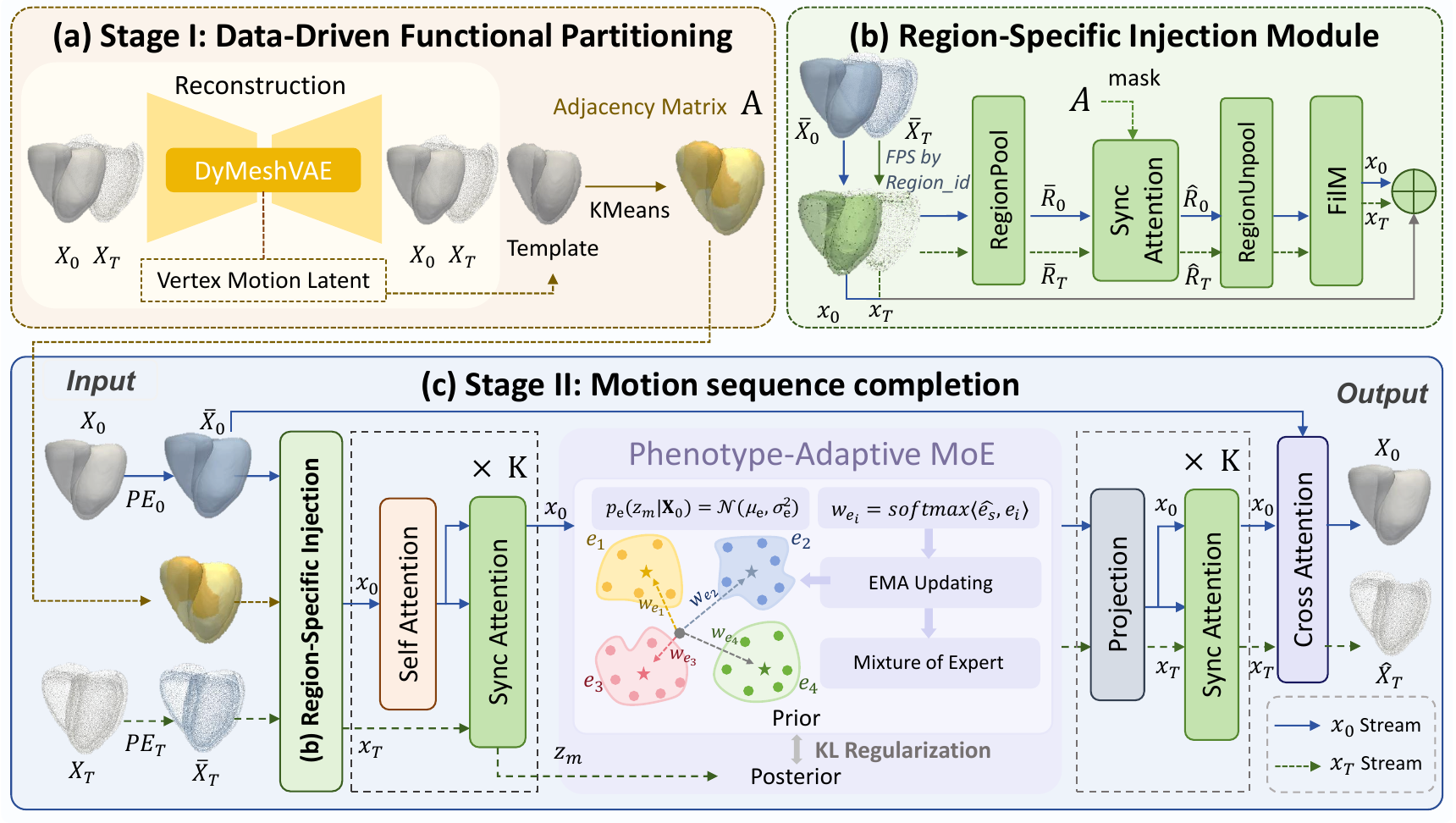}
  \caption{\textbf{Overview of the proposed RePCM framework.} (a) Stage I: Data-Driven Functional Partitioning. (b) Region-Specific Injection Module. (c) Stage II: Cardiac Motion Sequence Completion.}
  \label{fig:pipeline}
\end{figure*}

Our framework aims to learn multi-disease, region-aware cardiac dynamics and performs single-frame cardiac motion completion.
Given an end-diastolic (ED) mesh $\mathbf{X}_0 \in \mathbb{R}^{N \times 3}$, our goal is to synthesize a full cardiac cycle $\hat{\mathbf{X}}^{0:T-1} \in \mathbb{R}^{T \times N \times 3}$ by predicting ED-relative trajectories $\hat{\mathbf{X}}_{T} \in \mathbb{R}^{N \times (T\cdot 3)}$:
\begin{equation}
\hat{\mathbf{X}}^{t} = \mathbf{X}_0 + \hat{\mathbf{X}}_{T}^{t},\qquad t=0,\ldots,T-1.
\label{eq:task}
\end{equation}
As illustrated in Fig.~\ref{fig:pipeline}, the RePCM has two stages:
(I) Data-Driven Functional Partitioning: We derive a cohort-level partition to capture region-specific motion patterns, yielding a vertex-to-region assignment and region adjacency prior.
(II) Cardiac Motion sequence completion: we incorporate the region prior into a conditional VAE, where region-conditioned communication is enforced by the Region-Specific Injection Module, and inter-disease variability is modeled by a Phenotype-Adaptive MoE prior conditioned on $\mathbf{X}_0$.

\subsection{Stage I: Data-Driven Functional Partitioning}
We adopt DyMeshVAE~\cite{dymeshvae} as a motion feature extractor for functional partitioning.
The reconstructor encodes ED-relative trajectories $\mathbf{X}_{T}^{t}=\mathbf{X}^{t}-\mathbf{X}_{0}$ into a vertex-wise motion descriptor $\mathbf{z}^{(v)}\in\mathbb{R}^{d}$.
% 我们采用DyMeshVAE作为功能划分的运动特征提取器将ED 相对轨迹编码为逐顶点运动表示特征。
We then cluster $\{\mathbf{z}^{(v)}\}_{v=1}^{N}$ using KMeans to obtain a functional partition, assigning each vertex to one of $R$ motion regions.
With consistent vertex correspondence across subjects, this vertex-wise clustering yields a shared partition for the cohort. Based on these, we further build a binary region adjacency prior $\mathbf{A}\in\{0,1\}^{R\times R}$ to specify allowable inter-region communication in the completion model.

\subsection{Stage II: Cardiac Motion Sequence Completion}
%我们首先编码X0 in X0_ 通过raim得到位置感知后的锚点特征(encoder) 经过一个moe得到先验表示 最后decoder

\subsubsection{Region-Prior Injected Anchor Encoder}
Given the ED mesh $\mathbf{X}_0$ and relative trajectories $\mathbf{X}_T$, we obtain vertex features by applying distinct positional encodings to their coordinates, $\bar{\mathbf{X}}_0 = PE_0(\mathbf{X}_0)$ and $\bar{\mathbf{X}}_T = PE_T(\mathbf{X}_T)$, following~\cite{dymeshvae}. Farthest point sampling (FPS) then selects $K$ anchor vertices on the input ($K=512$), and their features are gathered as anchor tokens. Each anchor inherits the region ID of its sampled vertex.

\textbf{Region-Specific Injection Module}
To inject region priors into the completion encoder, we introduce a region-aware exchange module. Cardiac motion is highly heterogeneous across anatomical locations~\cite{LV,RV}, while global attention tends to over-smooth localized dynamics and mix functionally different parts. Therefore, the learned region prior is injected to explicitly constrain information exchange using the adjacency mask $\mathbf{A}$. Region tokens are obtained by pooling anchor tokens within each region, producing $\bar{R}_0$ and $\bar{R}_T$. Masked SyncAttention is applied at the region level: inter-region routing is computed from $x_0$, masked by $\mathbf{A}$ via Hadamard product, and reused to update both streams.
\begin{equation}
\begin{aligned}
\hat{R}_0&=\mathrm{Softmax}\!\left(\frac{\bar{R}_0\,\bar{R}_0^\top}{\sqrt{d_k}}\ \odot\ \mathbf{A}\right)\bar{R}_0+\bar{R}_0,\\
\hat{R}_T&=\mathrm{Softmax}\!\left(\frac{\bar{R}_0\,\bar{R}_0^\top}{\sqrt{d_k}}\ \odot\ \mathbf{A}\right)\bar{R}_T+\bar{R}_T,
\end{aligned}
\label{eq:raim_sync}
\end{equation}
where $d_k$ denotes the channel dimension of the projected space and $\odot$ is the Hadamard product. The updated region tokens $\hat{R}_0,\hat{R}_T$ are then broadcast back to anchors according to region IDs, providing region-aware context for anchor tokens $x_0$ and $x_T$. Finally, the injected region information modulates anchor features via FiLM~\cite{perez2018film} residual updates:
\begin{equation}
x_0 \leftarrow x_0 + \mathrm{FiLM}\!\left(\mathrm{LN}(x_0);\ \hat{R}_0\right),\qquad
x_T \leftarrow x_T + \mathrm{FiLM}\!\left(\mathrm{LN}(x_T);\ \hat{R}_T\right).
\label{eq:raim_film}
\end{equation}
This produces region-consistent, adjacency-constrained exchange while preserving stream-specific content, producing region-aware anchor representations for the trunk. A stack of attention layers further refines the anchor tokens. Each layer applies self-attention on $x_0$ and a SyncAttention block that reuses routing from $x_0$ to update both $x_0$ and $x_T$, producing the final anchor representations.

\subsubsection{Phenotype-Adaptive MoE}
During training, the encoder infers a motion posterior $q_\phi(\mathbf{z}_m \mid \mathbf{X}_{0:T-1})$ and regularizes it towards the shape-conditioned prior $p_\theta(\mathbf{z}_m \mid \mathbf{X}_0)$. A single disease label may not faithfully reflect motion phenotypes, whereas motion patterns are tied to shape. To capture disease-specific variability, we parameterize this prior with a shape-conditioned, phenotype-adaptive MoE. Specifically, we compute a global shape embedding $\mathbf{e}_s$ by pooling shape tokens $\mathbf{x}_0$ over anchors. We maintain $E$ prototypes $\{\mathbf{p}_e\}_{e=1}^{E}$ in the same embedding space and obtain soft expert weights by cosine-similarity assignment (no temperature). Each expert predicts Gaussian prior parameters $(\boldsymbol{\mu}_e,\boldsymbol{\sigma}_e^2)=h_e(\mathbf{x}_0)$, and we combine experts in parameter space to form the final prior:
\begin{equation}
\begin{aligned}
w_e &= \mathrm{Softmax}\!\big(\langle \hat{\mathbf{e}}_s,\hat{\mathbf{p}}_e\rangle \big), \quad e=1,\ldots,E,\\
\boldsymbol{\mu}_p &= \sum_{e=1}^{E} w_e\,\boldsymbol{\mu}_e,\qquad
\boldsymbol{\sigma}_p^2 = \sum_{e=1}^{E} w_e\,\boldsymbol{\sigma}_e^2,
\end{aligned}
\end{equation}
where $\hat{\cdot}$ denotes $\ell_2$-normalization. This defines $p_\theta(\mathbf{z}_m\mid \mathbf{X}_0)=\mathcal{N}(\boldsymbol{\mu}_p,\mathrm{diag}(\boldsymbol{\sigma}_p^2))$. The prototypes are updated using an exponential moving average of assigned embeddings, providing stable adaptive phenotype clustering during training.

\subsubsection{Anchor-Guided Trajectory Decoder}
The sampled latents are projected to anchor tokens $x_0$ and $x_T$ and refined by a stack of SyncAttention layers. We recover dense trajectories by a cross-attention write-back from vertices to anchors, using $x_0$ as keys and $x_T$ as values:
\begin{equation}
\mathbf{H}
=
\mathrm{Softmax}\!\left(\frac{\bar{\mathbf{X}}_0\,x_0^\top}{\sqrt{d_k}}\right)x_T.
\label{eq:vertex_writeback}
\end{equation}
The predicted trajectories are obtained by
\begin{equation}
\Delta\hat{\mathbf{X}}_T=\mathrm{Reshape}\!\left(\mathrm{Linear}(\mathbf{H})\right)\in\mathbb{R}^{T\times N\times 3}.
\label{eq:pred_traj}
\end{equation}

\subsection{Loss Function}

The objective is adapted from the evidence lower bound in \cite{elbo} and consists of two parts: 
\begin{equation}
\mathcal{L}=\mathcal{L}_{\mathrm{rec}}+\beta\,\mathcal{L}_{\mathrm{KL}}.
\label{eq:loss_all}
\end{equation}
The reconstruction loss $\mathcal{L}_{\mathrm{rec}}$ measures the discrepancy between the predicted mesh sequence and the ground-truth sequence. To encourage accurate estimation of motion patterns, we use an MSE loss over all vertices and time frames:
\begin{equation}
\mathcal{L}_{\mathrm{rec}}
=
\frac{1}{TN}\sum_{t=0}^{T-1}\sum_{v=1}^{N}
\left\|\hat{\mathbf{x}}_{t}^{(v)}-\mathbf{x}_{t}^{(v)}\right\|_2^2,
\label{eq:loss_rec}
\end{equation}
The KL loss $\mathcal{L}_{\mathrm{KL}}$ measures the divergence between the approximate posterior and the prior distribution:
\begin{equation}
\mathcal{L}_{\mathrm{KL}}
=
D_{\mathrm{KL}}\!\left(
q_\phi(\mathbf{z}_m\mid \mathbf{X}_{0:T-1})
\ \|\ 
p_\theta(\mathbf{z}_m\mid \mathbf{X}_0)
\right).
\label{eq:loss_kl}
\end{equation}

\section{Experiments and Results}
%现在需要修改的地方时datasets部分 写的时候很多不确定的表述 
\subsection{Datasets and Preprocessing}

For validation, we employed three publicly available cine cardiac MRI datasets: ACDC \cite{acdc}, M\&Ms \cite{mms}, and M\&Ms-2 \cite{mms2}.For each dataset, segmentation masks are converted into subject-specific bi-ventricular surface meshes over the full cardiac cycle by fitting an SSM~\cite{bai} to multi-phase contours, using global alignment, non-rigid refinement, and temporal smoothing to obtain anatomically consistent and reproducible meshes. Our study considers four cardiac phenotypes, normal (NOR), dilated cardiomyopathy (DCM), hypertrophic cardiomyopathy (HCM), and right-ventricular abnormality (RV), covering major variations in ventricular morphology and function with sufficient sample size for cross-dataset analysis. The cohort includes 666 subjects from ACDC (120), M\&Ms-2 (260), and M\&Ms (286), with 194 NOR, 187 DCM, 175 HCM, and 110 RV cases. All meshes are aligned to the template frame via center-of-mass matching and rigid registration applied to all frames, then normalized by a global scaling factor. Sequences are temporally resampled to 25 frames and split by patient into train/val/test with a 7:1:2 ratio.

\subsection{Implementation Details}
%好像需要加上重建实现细节 
%encoder那一块表达好奇怪 
The proposed framework is implemented in PyTorch. All experiments are conducted on a single NVIDIA RTX 4090 GPU (24\,GB memory). The encoder and decoder are each composed of 8 attention layers. During encoding, 512 latent tokens are sampled from the input mesh using FPS.
The model is optimized using the AdamW optimizer with an initial learning rate of $1\times10^{-4}$ and a batch size of 8. The latent dimension is set to 16. Training is performed for up to 500 epochs with early stopping based on the validation performance. 

\subsection{Evaluation Metrics}
We evaluate BiV sequence completion using $\mathrm{ASSD}$, $\mathrm{HD95}$, and vertex-wise $\mathrm{RMSE}$, reported separately for LV and RV. All methods condition on the end-diastolic first-frame mesh $X_0$ to predict the full cardiac cycle. We compare with a conditional VAE and adapted baselines, including ACTOR~\cite{actor}, Action2Motion~\cite{a2m}, CHeart~\cite{CHEART}, MeshHeart~\cite{MeshHeart}, and CardioSynth4D~\cite{Cadio4d}, all evaluated under the same unified-topology BiV sequence completion setting.

\subsection{Results}
\subsubsection{Comparison Results}

\begin{figure*}[t!]
  \centering
  \includegraphics[width=.95\textwidth]{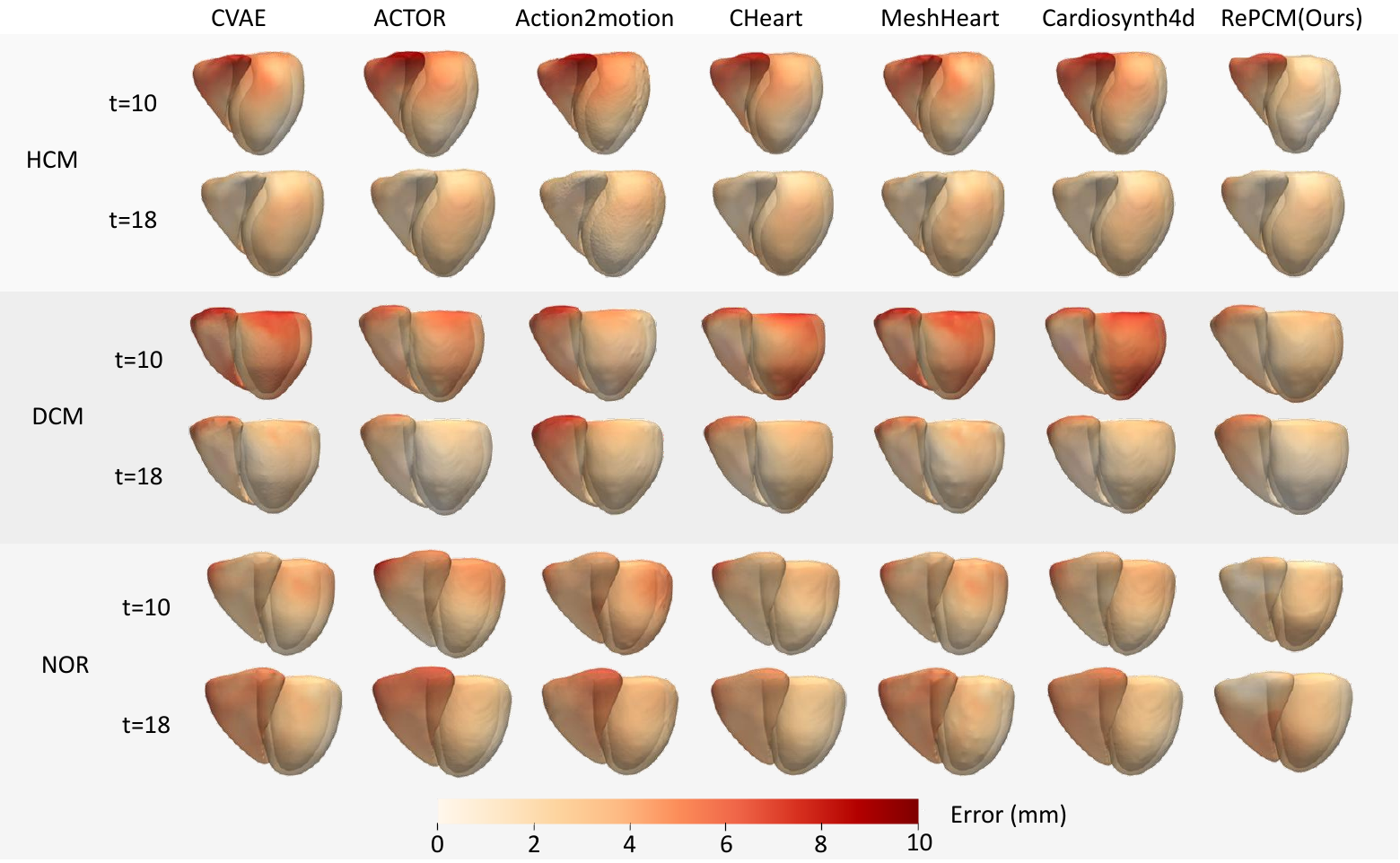}
  \caption{\textbf{Illustration of biventricular motion completion results.}
  HCM: hypertrophic cardiomyopathy.
  DCM: dilated cardiomyopathy.
  NOR: normal.}
  \label{fig:qualitative_disease}
\end{figure*}

\begin{figure*}[t!]
  \centering
  \includegraphics[width=.95\textwidth]{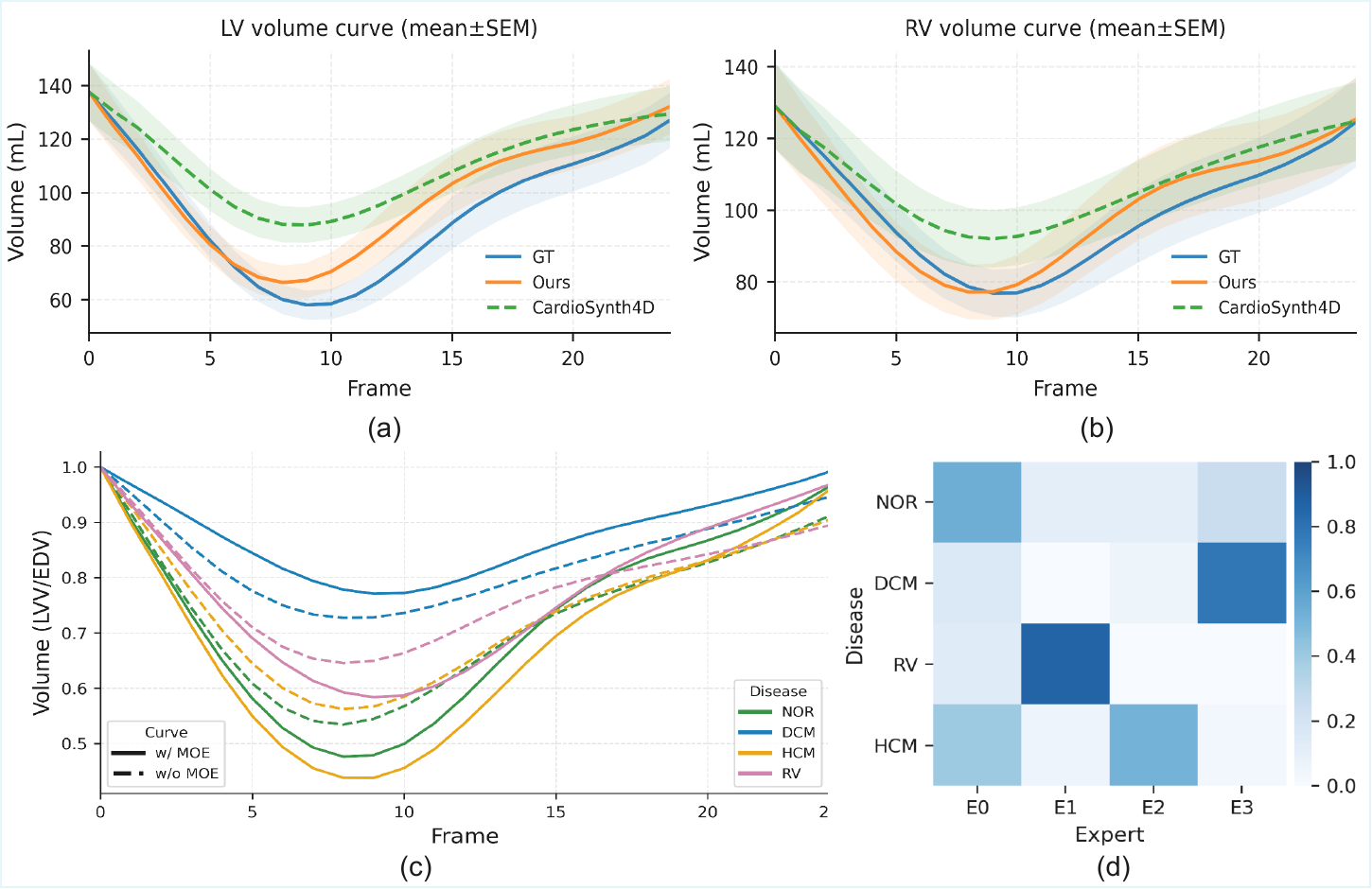}
  \caption{\textbf{Comparsion Results: }(a) LV volume curves. (b) RV volume curves. \textbf{Ablation Results: }(c) Normalized LVV/EDV curves by disease. (d) Expert-disease usage matrix showing  expert specialization.}
  \label{fig:volume_curves_lv_rv}
\end{figure*}

As summarized in Table~\ref{tab:cmp_lv_rv_meanstd_8pt}, our method achieves the best overall reconstruction accuracy across ASSD, HD95, and vtxRMSE. The largest gain is observed on the LV, where HD95 drops from 4.22 to 3.94~mm compared with CardioSynth4D, suggesting that our model more effectively suppresses localized failure regions and preserves fine-grained LV geometry. Fig.~\ref{fig:qualitative_disease} visualizes three representative cases at two time points (t{=}10, t{=}18). Across diseases and phases, our predictions exhibit the smallest error regions and better preserve fine-grained ventricular morphology. Notably, RV remains challenging across all methods due to complex anatomy and larger inter-subject variability, but our model reduces both average and near-worst-case RV errors. 
Consistent with these findings, Fig.~\ref{fig:volume_curves_lv_rv} (a) shows that our method follows the reference volume trajectories more closely throughout the cardiac cycle with a narrower uncertainty band, suggesting more temporally consistent dynamics.

\begin{table}[t!]
\centering
\begingroup
\fontsize{8}{9.6}\selectfont
\setlength{\tabcolsep}{2.6pt}
\renewcommand{\arraystretch}{1.08}
\caption{Comparison on LV/RV surface reconstruction accuracy (mean$\pm$std).}
\label{tab:cmp_lv_rv_meanstd_8pt}
\resizebox{\linewidth}{!}{%
\begin{tabular}{lcccccc}
\toprule
Methods
& \multicolumn{2}{c}{ASSD (mm)$\downarrow$}
& \multicolumn{2}{c}{HD95 (mm)$\downarrow$}
& \multicolumn{2}{c}{vtxRMSE (mm)$\downarrow$} \\
\cmidrule(lr){2-3}\cmidrule(lr){4-5}\cmidrule(lr){6-7}
& LV & RV & LV & RV & LV & RV \\
\midrule
CVAE          & $2.16\pm0.66$ & $2.08\pm0.63$ & $4.34\pm1.28$ & $4.80\pm1.75$ & $3.86\pm1.32$ & $4.25\pm1.52$ \\
ACTOR         & $2.10\pm0.63$ & $2.20\pm0.79$ & $4.28\pm1.39$ & $4.85\pm2.30$ & $3.71\pm1.38$ & $4.16\pm1.81$ \\
Action2Motion & $2.28\pm0.70$ & $2.18\pm0.77$ & $4.81\pm1.63$ & $5.07\pm2.15$ & $4.20\pm1.51$ & $4.51\pm1.83$ \\
CHeart        & $2.21\pm0.60$ & \underline{$2.04\pm0.64$} & $4.48\pm1.27$ & \underline{$4.73\pm1.70$} & $3.80\pm1.23$ & \underline{$4.14\pm1.49$} \\
MeshHeart     & $2.09\pm0.64$ & $2.11\pm0.62$ & $4.23\pm1.28$ & $4.78\pm1.75$ & $3.78\pm1.32$ & $4.20\pm1.52$ \\
CardioSynth4D & \underline{$2.07\pm0.64$} & $2.04\pm0.71$ & \underline{$4.22\pm1.41$} & $4.73\pm1.95$ & \underline{$3.63\pm1.36$} & $4.16\pm1.68$ \\
\midrule
RePCM (Ours)  & $\mathbf{1.79\pm0.60}$ & $\mathbf{2.00\pm0.76}$ & $\mathbf{3.94\pm1.48}$ & $\mathbf{4.54\pm2.04}$ & $\mathbf{3.43\pm1.42}$ & $\mathbf{4.08\pm1.74}$ \\
\bottomrule
\end{tabular}%
}
\endgroup
\end{table}

\begin{table}[t!]
\centering
\begingroup
\fontsize{8}{9.6}\selectfont
\setlength{\tabcolsep}{2.4pt}
\renewcommand{\arraystretch}{1.08}
\caption{Ablation on the region partition $R$ and expert number $E$. }
\label{tab:ablation_rk_meanstd_8pt}
\resizebox{\linewidth}{!}{%
\begin{tabular}{llcccccc}
\toprule
\multirow{2}{*}{MoE} & \multirow{2}{*}{Setting}
& \multicolumn{2}{c}{ASSD (mm) $\downarrow$}
& \multicolumn{2}{c}{HD95 (mm) $\downarrow$}
& \multicolumn{2}{c}{vtxRMSE (mm) $\downarrow$} \\
\cmidrule(lr){3-4}\cmidrule(lr){5-6}\cmidrule(lr){7-8}
& & LV & RV & LV & RV & LV & RV \\
\midrule
\multirow{3}{*}{w/o MoE}
& R=16 & $1.90\pm0.69$ & ${2.02\pm0.80}$ & $4.05\pm1.50$ & $4.80\pm2.22$ & $3.51\pm1.46$ & $4.17\pm1.83$ \\
& R=24 & ${1.98\pm0.59}$ & $2.06\pm0.77$ & ${4.16\pm1.46}$ & ${4.70\pm2.09}$ & ${3.56\pm1.41}$ & ${4.17\pm1.82}$ \\
& R=32 & $1.97\pm0.60$ & $2.04\pm0.72$ & $4.15\pm1.45$ & $4.91\pm2.04$ & $3.59\pm1.43$ & $4.18\pm1.72$ \\
\midrule
\multirow{3}{*}{w/ MoE}
& E=4 & ${1.79\pm0.60}$ & ${2.00\pm0.76}$ & ${3.94\pm1.48}$ & ${4.54\pm2.04}$ & ${3.43\pm1.42}$ & ${4.08\pm1.74}$ \\
& E=6 & $1.90\pm0.60$ & $2.07\pm0.71$ & $4.07\pm1.41$ & $4.74\pm1.94$ & $3.55\pm1.34$ & $4.19\pm1.63$ \\
& E=8 & $1.86\pm0.60$ & $2.03\pm0.74$ & $4.12\pm1.50$ & $4.80\pm2.09$ & $3.58\pm1.43$ & $4.19\pm1.78$ \\
\bottomrule
\end{tabular}%
}
\endgroup
\end{table}

\subsubsection{Ablation Results}
Table~\ref{tab:ablation_rk_meanstd_8pt} studies the number of motion regions $R$ and experts $E$. The best BiV generation is achieved with $R{=}16$, while larger $R$ slightly degrades performance, suggesting over-partitioning weakens region-specific motion cues and encourages more averaged global dynamics. Adding MoE improves our base framework with the best setting at $E{=}4$, whereas larger $E$ reduces accuracy due to fewer samples per expert and less stable routing. 
Fig.~\ref{fig:volume_curves_lv_rv} (c) shows disease-wise LV volume trajectories normalized by EDV. Compared with the non-MoE variant, the MoE-enhanced model produces more phenotype-distinct dynamics, suggesting improved capture of phenotype-dependent motion patterns. Fig.~\ref{fig:volume_curves_lv_rv} (d) further visualizes phenotype-aware routing by assigning each test case to the expert with the highest gating weight. The resulting usage matrix shows clearer expert preference for DCM and RV, while NOR and HCM exhibit more mixed routing, consistent with their partially overlapping motion characteristics.

% \begin{figure}[!htbp]
%   \centering
%   \includegraphics[width=.95\textwidth]{fig3.pdf}
%   \caption{\textbf{Ventricular volume curves (mean$\pm$SEM).}
%   Left-ventricular (LV) and right-ventricular (RV) volume curves over the cardiac cycle comparing Ground Truth (GT), our method, and CardioSynth4D.}
%   \label{fig:volume_curves_lv_rv}
% \end{figure}

% \begin{figure*}[!htbp]
%   \centering
%   \includegraphics[width=.95\textwidth]{fig4.pdf}
%   \caption{\textbf{Effect of Phenotype-Adaptive MoE.}
%   {(a) Normalized LV volume curves (LVV/EDV) across diseases comparing w\/ MoE vs. w\/o MoE}. (b) Expert - Diseaseusage matrix illustrating how experts specialize to different phenotypes.}
%   \label{fig:moe_ablation}
% \end{figure*}

\section{Conclusion}
We propose a multi-disease, region-aware conditional generative framework for single-frame biventricular mesh motion completion. The framework links cardiac shape and motion across disease populations through a two-stage design. First, reconstruction motion descriptors are clustered to obtain cohort-level motion regions and a region adjacency prior. Second, the priors are injected into a conditional VAE through a Region-Specific Injection Module, promoting region-consistent information exchange. A shape-conditioned Phenotype-Adaptive MoE prior captures disease-specific variability. Experiments on three public datasets show improved motion fidelity and better preservation of region-specific dynamics over state-of-the-art methods. Future work will extend to virtual population generation with broader phenotypes and whole-heart anatomy.

\begin{credits}
\subsubsection{\ackname} This work was supported by Singapore National Medical Research Council Open Fund - Young Individual Research Grant (25-1321-A0001), and Singapore Ministry of Education Tier 1 grant (25-1097-P0001).

\subsubsection{\discintname}
The authors have no competing interests to declare that are relevant to the content of this article.
\end{credits}

\bibliographystyle{splncs04}
\bibliography{references}

@misc{intro[1],
  title={Myocardial strain imaging: theory, current practice, and the future. JACC Cardiovasc Imaging},
  author={Smiseth, OA and Rider, O and Cvijic, M and Valkovic, L and Remme, EW and Voigt, JU},
  year={2024}
}

@article{intro[2],
  title={Characterizing interactions between cardiac shape and deformation by non-linear manifold learning},
  author={Di Folco, Maxime and Moceri, Pamela and Clarysse, Patrick and Duchateau, Nicolas},
  journal={Medical image analysis},
  volume={75},
  pages={102278},
  year={2022},
  publisher={Elsevier}
}

@article{intro[3],
  title={Myocardial strain imaging: theory, current practice, and the future},
  author={Smiseth, Otto A and Rider, Oliver and Cvijic, Marta and Valkovi{\v{c}}, Ladislav and Remme, Espen W and Voigt, Jens-Uwe},
  journal={JACC: Cardiovascular Imaging},
  volume={18},
  number={3},
  pages={340--381},
  year={2025},
  publisher={Elsevier}
}

@article{intro[4],
  title={Retrospective electrocardiography-gated real-time cardiac cine MRI at 3T: comparison with conventional segmented cine MRI},
  author={Cui, Chen and Yin, Gang and Lu, Minjie and Chen, Xiuyu and Cheng, Sainan and Li, Lu and Yan, Weipeng and Song, Yanyan and Prasad, Sanjay and Zhang, Yan and others},
  journal={Korean journal of radiology},
  volume={20},
  number={1},
  pages={114--125},
  year={2019},
  publisher={The Korean Society of Radiology}
}

@inproceedings{SSM,
  title={A statistical shape model of the heart and its application to model-based segmentation},
  author={Ordas, Sebastian and Oubel, Estanislao and Leta, Rub{\'e}n and Carreras, Francesc and Frangi, Alejandro F},
  booktitle={Medical Imaging 2007: Physiology, Function, and Structure from Medical Images},
  volume={6511},
  pages={490--500},
  year={2007},
  organization={SPIE}
}

@article{SDF4CHD,
  title={SDF4CHD: Generative modeling of cardiac anatomies with congenital heart defects},
  author={Kong, Fanwei and Stocker, Sascha and Choi, Perry S and Ma, Michael and Ennis, Daniel B and Marsden, Alison L},
  journal={Medical image analysis},
  volume={97},
  pages={103293},
  year={2024},
  publisher={Elsevier}
}

@inproceedings{STNDF,
  title={Spatio-temporal neural distance fields for conditional generative modeling of the heart},
  author={S{\o}rensen, Kristine and Diez, Paula and Margeta, Jan and El Youssef, Yasmin and Pham, Michael and Pedersen, Jonas Jalili and K{\"u}hl, Tobias and De Backer, Ole and Kofoed, Klaus and Camara, Oscar and others},
  booktitle={International Conference on Medical Image Computing and Computer-Assisted Intervention},
  pages={422--432},
  year={2024},
  organization={Springer}
}

@article{CHEART,
  title={Cheart: A conditional spatio-temporal generative model for cardiac anatomy},
  author={Qiao, Mengyun and Wang, Shuo and Qiu, Huaqi and De Marvao, Antonio and O’Regan, Declan P and Rueckert, Daniel and Bai, Wenjia},
  journal={IEEE transactions on medical imaging},
  volume={43},
  number={3},
  pages={1259--1269},
  year={2023},
  publisher={IEEE}
}

@article{MeshHeart,
  title={A personalized time-resolved 3D mesh generative model for unveiling normal heart dynamics},
  author={Qiao, Mengyun and McGurk, Kathryn A and Wang, Shuo and Matthews, Paul M and O’Regan, Declan P and Bai, Wenjia},
  journal={Nature Machine Intelligence},
  volume={7},
  number={5},
  pages={800--811},
  year={2025},
  publisher={Nature Publishing Group UK London}
}

@inproceedings{Cadio4d,
  title={4D CardioSynth: Synthesising Dynamic Virtual Heart Populations Through Spatiotemporal Disentanglement},
  author={Dou, Haoran and Huang, Jinghan and Zakeri, Arezoo and Zhou, Zherui and Mu, Tingting and Duan, Jinming and Frangi, Alejandro F},
  booktitle={International Conference on Medical Image Computing and Computer-Assisted Intervention},
  pages={3--12},
  year={2025},
  organization={Springer}
}

@inproceedings{dymeshvae,
  title={Animateanymesh: A feed-forward 4d foundation model for text-driven universal mesh animation},
  author={Wu, Zijie and Yu, Chaohui and Wang, Fan and Bai, Xiang},
  booktitle={Proceedings of the IEEE/CVF International Conference on Computer Vision},
  pages={13557--13568},
  year={2025}
}

@article{elbo,
  title={Auto-encoding variational bayes},
  author={Kingma, Diederik P and Welling, Max},
  journal={arXiv preprint arXiv:1312.6114},
  year={2013}
}

@inproceedings{perez2018film,
  title={Film: Visual reasoning with a general conditioning layer},
  author={Perez, Ethan and Strub, Florian and De Vries, Harm and Dumoulin, Vincent and Courville, Aaron},
  booktitle={Proceedings of the AAAI conference on artificial intelligence},
  volume={32},
  number={1},
  year={2018}
}

@article{RV,
  title={Sex-and method-specific reference values for right ventricular strain by 2-dimensional speckle-tracking echocardiography},
  author={Muraru, Denisa and Onciul, Sebastian and Peluso, Diletta and Soriani, Nicola and Cucchini, Umberto and Aruta, Patrizia and Romeo, Gabriella and Cavalli, Giacomo and Iliceto, Sabino and Badano, Luigi P},
  journal={Circulation: Cardiovascular Imaging},
  volume={9},
  number={2},
  pages={e003866},
  year={2016},
  publisher={Lippincott Williams \& Wilkins Hagerstown, MD}
}

@article{LV,
  title={Transmural gradients of myocardial structure and mechanics: implications for fiber stress and strain in pressure overload},
  author={Carruth, Eric D and McCulloch, Andrew D and Omens, Jeffrey H},
  journal={Progress in biophysics and molecular biology},
  volume={122},
  number={3},
  pages={215--226},
  year={2016},
  publisher={Elsevier}
}

@article{acdc,
  title={Deep learning techniques for automatic MRI cardiac multi-structures segmentation and diagnosis: is the problem solved?},
  author={Bernard, Olivier and Lalande, Alain and Zotti, Clement and Cervenansky, Frederick and Yang, Xin and Heng, Pheng-Ann and Cetin, Irem and Lekadir, Karim and Camara, Oscar and Ballester, Miguel Angel Gonzalez and others},
  journal={IEEE transactions on medical imaging},
  volume={37},
  number={11},
  pages={2514--2525},
  year={2018},
  publisher={ieee}
}

@article{mms,
  title={Multi-centre, multi-vendor and multi-disease cardiac segmentation: the M\&Ms challenge},
  author={Campello, Victor M and Gkontra, Polyxeni and Izquierdo, Cristian and Martin-Isla, Carlos and Sojoudi, Alireza and Full, Peter M and Maier-Hein, Klaus and Zhang, Yao and He, Zhiqiang and Ma, Jun and others},
  journal={IEEE Transactions on Medical Imaging},
  volume={40},
  number={12},
  pages={3543--3554},
  year={2021},
  publisher={IEEE}
}

@article{mms2,
  title={Deep learning segmentation of the right ventricle in cardiac MRI: the M\&Ms challenge},
  author={Mart{\'\i}n-Isla, Carlos and Campello, V{\'\i}ctor M and Izquierdo, Cristian and Kushibar, Kaisar and Sendra-Balcells, Carla and Gkontra, Polyxeni and Sojoudi, Alireza and Fulton, Mitchell J and Arega, Tewodros Weldebirhan and Punithakumar, Kumaradevan and others},
  journal={IEEE Journal of Biomedical and Health Informatics},
  volume={27},
  number={7},
  pages={3302--3313},
  year={2023},
  publisher={IEEE}
}

@inproceedings{a2m,
  title={Action2motion: Conditioned generation of 3d human motions},
  author={Guo, Chuan and Zuo, Xinxin and Wang, Sen and Zou, Shihao and Sun, Qingyao and Deng, Annan and Gong, Minglun and Cheng, Li},
  booktitle={Proceedings of the 28th ACM international conference on multimedia},
  pages={2021--2029},
  year={2020}
}

@inproceedings{actor,
  title={Action-conditioned 3d human motion synthesis with transformer vae},
  author={Petrovich, Mathis and Black, Michael J and Varol, G{\"u}l},
  booktitle={Proceedings of the IEEE/CVF international conference on computer vision},
  pages={10985--10995},
  year={2021}
}

@article{bai,
  title={A bi-ventricular cardiac atlas built from 1000+ high resolution MR images of healthy subjects and an analysis of shape and motion},
  author={Bai, Wenjia and Shi, Wenzhe and de Marvao, Antonio and Dawes, Timothy JW and O’Regan, Declan P and Cook, Stuart A and Rueckert, Daniel},
  journal={Medical image analysis},
  volume={26},
  number={1},
  pages={133--145},
  year={2015},
  publisher={Elsevier}
}
\end{document}